%% file: neural-fixed-point.tex
\documentclass[twoside,11pt]{article}

\usepackage[nohyperref]{automl2021}
\input{math_commands.tex}

\usepackage{microtype}
\usepackage{graphicx}
\usepackage{subcaption}
\usepackage{booktabs} 

\usepackage{xcolor}
\definecolor{linkcolor}{RGB}{168, 141, 201}
\usepackage[colorlinks=true,allcolors=linkcolor,pageanchor=true,plainpages=false,pdfpagelabels,bookmarks,bookmarksnumbered]{hyperref}

\usepackage{algorithm}

\usepackage{algpseudocode}
\algnewcommand{\LeftCommentX}[1]{\Statex \(\triangleright\) #1}
\algnewcommand{\LeftComment}[1]{\State \(\triangleright\) #1}

\usepackage{tikz}
\newcommand{\cblock}[3]{
  \hspace{-1.5mm}
  \begin{tikzpicture}
    [
    node/.style={square, minimum size=10mm, thick, line width=0pt},
    ]
    \node[fill={rgb,255:red,#1;green,#2;blue,#3}] () [] {};
  \end{tikzpicture}%
}

\makeatletter
\DeclareRobustCommand\onedot{\futurelet\@let@token\@onedot}
\def\@onedot{\ifx\@let@token.\else.\null\fi\xspace}
\newcommand{\eg}{\emph{e.g}\onedot}
\newcommand{\ie}{\emph{i.e}\onedot}
\makeatother

\usepackage[nameinlink]{cleveref}
\Crefname{algorithm}{Alg.}{Algs.}
\Crefname{section}{Sect.}{Sects.}
\Crefname{appendix}{App.}{Apps.}
\Crefname{proposition}{Prop.}{Props.}

\usepackage{xcolor}
\definecolor{linkcolor}{RGB}{74, 102, 146}
\definecolor{light_purple}{RGB}{168, 141, 201}

\usepackage{todonotes}

\usepackage{enumitem}
\setlist[enumerate]{topsep=0pt,itemsep=-1ex,partopsep=1ex,parsep=1ex}
\setlist[itemize]{topsep=0pt,itemsep=-1ex,partopsep=1ex,parsep=1ex}

\newcommand{\initnet}{\ensuremath{g^{\rm init}_\theta}\xspace}
\newcommand{\accnet}{\ensuremath{g^{\rm acc}_\theta}\xspace}

\jmlrheading{S.~Venkataraman and B.~Amos}
\ShortHeadings{Neural Fixed-Point Acceleration}{Venkataraman and Amos}
\firstpageno{1}

\begin{document}

\title{Neural Fixed-Point Acceleration for Convex Optimization}
\author{\name Shobha Venkataraman$^*$ \\ \name Brandon Amos\thanks{Equal contribution.} \\ \addr Facebook AI}
\maketitle

\begin{abstract}
  Fixed-point iterations are at the heart of numerical computing and
  are often a computational bottleneck in real-time applications
  that typically need a fast solution of moderate accuracy.
  We present \emph{neural fixed-point acceleration}
  which combines ideas from meta-learning and classical acceleration
  methods to automatically learn to accelerate fixed-point problems
  that are drawn from a distribution.
  We apply our framework to SCS, the state-of-the-art solver for convex
  cone programming, and design
  models and loss functions to overcome the challenges of learning
  over unrolled optimization and acceleration instabilities.
  Our work brings neural acceleration into any optimization problem expressible
  with CVXPY.
  The source code behind this paper is available at
\href{https://github.com/facebookresearch/neural-scs}{github.com/facebookresearch/neural-scs}.
\end{abstract}

\section{Introduction}
Continuous fixed-point problems are a computational primitive
in numerical computing, optimization, machine learning, and
the natural and social sciences.
Given a map $f:\R^n\rightarrow\R^n$, a \emph{fixed point}
$x\in\R^n$ is where $f(x) = x$.
\emph{Fixed-point iterations} repeatedly apply $f$ until
the solution is reached and provably converge under
assumptions of $f$.
Most solutions to optimization problems can be seen as finding
a fixed point mapping of the iterates, \eg in the convex setting,
$f$ could step a primal-dual iterate closer to the
KKT optimality conditions of the problem,
which remains fixed once it is reached.
Recently in the machine learning community, fixed point computations
have been brought into the modeling pipeline through the use of
differentiable convex optimization
\citep{domke2012generic,gould2016differentiating,amos2017optnet,agrawal2019differentiable,lee2019meta},
differentiable control \citep{amos2018dmpc},
deep equilibrium models \citep{bai2019deep,bai2020multiscale},
and
sinkhorn iterations \citep{mena2018learning}.

Fixed-point computations are often a computational bottleneck in the
larger systems they are a part of. \emph{Accelerating} (\ie speeding
up) fixed point computations is an active area of optimization
research that involves using the knowledge of prior iterates to
improve the future ones.  These improve over standard fixed-point
iterations but are classically done without learning. The optimization
community has traditionally not explored learned solvers because of
the lack of theoretical guarantees on learned solvers.
For many real-time applications, though, traditional
fixed-point solvers can be too slow; instead we need
a fast low-accuracy solution. Further, fixed-point problems
repeatedly solved in an application typically share a lot of structure
and so an application naturally induces a distribution of fixed-point
{\em problem instances}. This raises the question: can we
learn a fast and sufficiently-accurate fixed-point solver, when the
problem instances are drawn from a fixed distribution?

\begin{figure}[t]
  \vspace{-5mm}
  \centering
  \begin{subfigure}{0.43\textwidth}
      \includegraphics[width=0.99\linewidth]{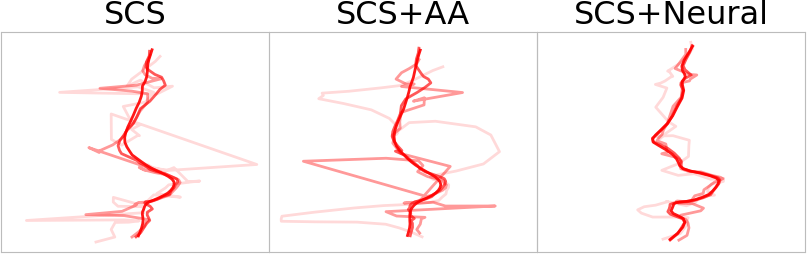}
      \includegraphics[width=0.99\linewidth]{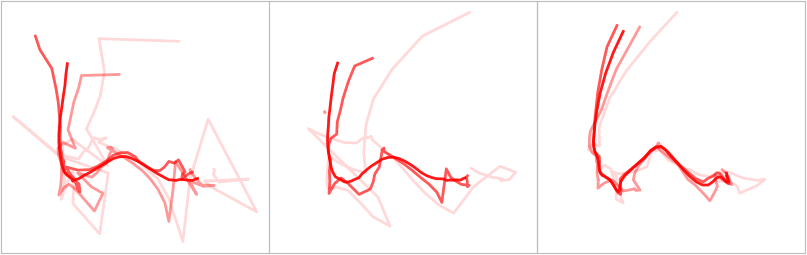}
      \caption{Robust Kalman filtering: Solutions at iterations 5, 10,
        20 \& 50. Colors get progressively darker at higher
        iterations.}
      \label{fig:kalman_samples}
  \end{subfigure}
\begin{subfigure}{0.5\textwidth}
  \centering
    \includegraphics[width=0.98\linewidth]{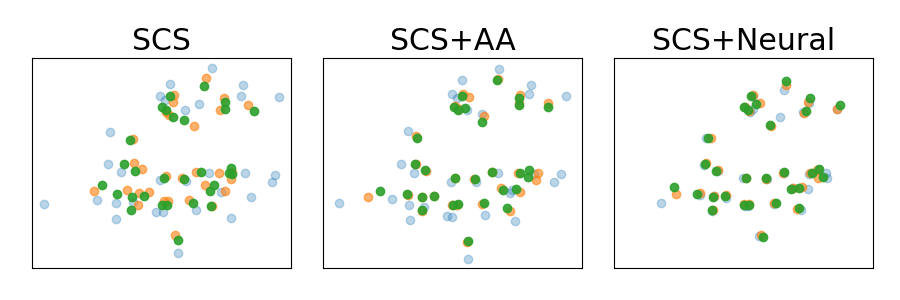}
    \includegraphics[width=0.96\linewidth]{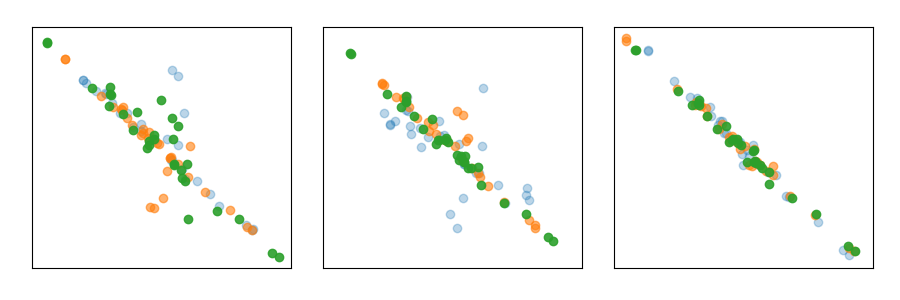}
  \caption{Robust PCA, iteration:
  \cblock{187}{214}{232}\hspace{1.5mm}10\hspace{1.5mm}
  \cblock{255}{178}{110}\hspace{1.5mm}15\hspace{1.5mm}
  \cblock{44}{159}{44}\hspace{1.5mm}20
  }
  \label{fig:pca_samples}
\end{subfigure}
\label{fig:visual}
\caption{Visualizing neural-accelerated test instances for robust
  Kalman filtering and robust PCA. Each line is a single instance.
  Neural-accelerated SCS quickly stabilizes to a solution, while the
  SCS and SCS+AA iterations exhibit higher variance.}
\end{figure}

In this paper, we explore the problem of learning to accelerate
fixed-point problem instances drawn from a distribution, which we term
\emph{neural fixed-point acceleration}. We focus on convex
optimization to ground our work in real applications,
including real-time ones such as \citet{tedrake2010lqr,mattingley2010real}.
We design a
framework for our problem based on {\em learning to optimize}, \ie,
meta-learning (\Cref{sec:related}): we learn a model that
accelerates the fixed-point computations on a fixed distribution of
problems, by repeatedly backpropagating through their unrolled
computations. We build on ideas from classical acceleration:
we learn a model that uses the prior iterates to
improve them, for problems in this
distribution. Our framework also captures classical acceleration
methods as an instance.

We show how we can learn an acceleration model for convex cone
programming with this framework.  We focus on SCS \citep{o2016conic},
which is the state-of-the-art default cone solver in CVXPY
\citep{diamond2016cvxpy}. However, learning to optimize and
acceleration are notoriously hard problems with instabilities and poor
solutions, so there are challenges in applying our framework to SCS,
which has complex fixed-point computations and
interdependencies. Through careful design of models and loss
functions, we address the challenges of differentiating through
unrolled SCS computations and the subtleties of interweaving model
updates with iterate history. Our experiments show that we
consistently accelerate SCS in three applications -- lasso, robust PCA
and robust Kalman filtering.

\section{Related work}\label{sec:related}
\paragraph{Learned optimizers and meta-learning.}
The machine learning community has recently explored many approaches
to learning to improve the solutions to optimization problems.
These applications have wide-ranging applications, \eg in optimal
power flow \citep{baker2020learning,donti2021dc}, combinatorial
optimization
\cite{khalil2016learning,dai2017learning,nair2020solving,bengio2020machine},
and differential equations
\citep{li2020fourier,poli2020hypersolvers,kochkov2021machine}.  The
meta-learning and learning to optimize literature, \eg
\citep{li2016learning,finn2017model,wichrowska2017learned,andrychowicz2016learning,metz2019using,metz2021training,gregor2010learning},
focuses on learning better solutions to parameter learning problems
that arise for machine learning tasks.
\citet{bastianello2021opreg} approximates the fixed-point iteration
with the closest contractive fixed-point iteration.
\citet{ichnowski2021accelerating} use reinforcement learning to
improve quadratic programming solvers.
Our work is the most strongly
connected to the learning to optimize work here and is an
application of these methods to fixed-point computations and convex
cone programming.

\paragraph{Fixed-point problems and acceleration.}
Accelerating fixed-point
computations date back decades and include
\emph{Anderson Acceleration} (AA) \citep{anderson1965iterative} and
\emph{Broyden's method} \citep{broyden1965class}, or
variations such as \citet{walker2011anderson,zhang2020globally}.

\section{Neural fixed-point acceleration}\label{sec:neural_fp}

\begin{algorithm}[t]
  \caption{Neural fixed-point acceleration augments standard
    fixed-point computations with a learned initialization
    and updates to the iterates.
  }
\label{alg:neural-fp}
\begin{algorithmic}
  \State \textbf{Inputs:} Context $\phi$, parameters $\theta$, and fixed-point map $f$.
  \State $[x_1, h_1] = \initnet(\phi)$
  \Comment Initial hidden state and iterate
  \For{fixed-point iteration $t = 1..T$}
  \State $\tilde x_{t+1}= f(x_{t}; \phi)$
  \Comment Original fixed-point iteration
    \State $x_{t+1}, h_{t+1} = \accnet(x_t, \tilde x_{t+1}, h_t)$
        \Comment Acceleration
  \EndFor
\end{algorithmic}
\end{algorithm}

\subsection{Problem formulation}
We are interested in settings and systems that involve
solving a known distribution over fixed-point problems.
Each fixed-point problem depends on
a \emph{context} $\phi\in\R^m$
that we have a distribution over $\gP(\phi)$.
The distribution $\gP(\phi)$ induces a distribution over
fixed-point problems $f(x; \phi)=x$ with a fixed-point
map $f$ that depends on the context.
Informally, our objective will be to solve this class
of fixed-point problems as fast as possible.
Notationally, other settings refer to $\phi$ as
a ``parameter'' or ``conditioning variable,''
but here we will consistently use ``context.''
We next consider a general solver for fixed-point problems that
captures classical acceleration methods as an instance,
and can also be parameterized with some $\theta$ and learned
to go beyond classical solvers.
Given a fixed context $\phi$, we solve the fixed-point problem
with \cref{alg:neural-fp}.
At each time step $t$ we maintain the \emph{fixed-point iterations} $x_t$
and a \emph{hidden state} $h_t$.
The \emph{initializer} \initnet depends on the context $\phi$ provides
the starting iterate and hidden state and the
\emph{acceleration} \accnet updates the iterate after observing
the application of the fixed-point map $f$.

\begin{proposition}
  \Cref{alg:neural-fp} captures Anderson Acceleration
  as stated \eg, in \citet{zhang2020globally}.
\end{proposition}

This can be seen by making the hidden state a list
of the previous $k$ fixed-point iterations, and there would be
no parameters $\theta$.
The initializer \initnet would return a deterministic, problem-specific
initial iterate, and the acceleration \accnet would apply the
standard update and append the fixed-point iteration
to the hidden state.

\subsection{Modeling and optimization}

We first parameterize the models behind the fixed-point updates in
\Cref{alg:neural-fp}.  In neural acceleration, we will use learned
models for \initnet and \accnet.
We experimentally found that we achieve good results a standard MLP for
\initnet and a recurrent model such as an LSTM
\citep{hochreiter1997long} or GRU \citep{cho2014learning} for
\accnet. While the appropriate models vary by application, a
recurrent structure is a particularly good fit as it
encapsulates the history of iterates in the hidden
state, and uses that to predict a future iterate.

Next, we define and optimize an objective for learning that
characterizes how well the fixed-point iterations are solved.  Here,
we use the \emph{fixed-point residual norms} defined by
$\gR(x; \phi)\defeq ||x-f(x; \phi)||_2$.
This is a natural choice for the objective as the convergence analysis of classical
acceleration methods are built around the fixed-point residual. Our
learning objective is thus to find the parameters to minimize the
fixed-point residual norms in every iteration across the distribution
of fixed-point problem instances, \ie
\begin{equation}
  \minimize_\theta\; \E_{\phi\sim \gP(\phi)} \sum_{t<T} \gR(x_t; \phi) / \gR_0(\phi),
  \label{eq:obj}
\end{equation}
where $T$ is the maximum number of iterations to apply and
$\gR_0$ is a normalization factor that is useful
when the fixed-point residuals have different magnitudes.
We optimize \cref{eq:obj} with gradient descent, which
requires the derivatives of the fixed-point map $\nabla_x f(x)$.

\section{Accelerating Convex Cone Programming}
\label{sec:cone_prog}
We have added neural acceleration to SCS (\emph{Neural SCS}) and
integrated it with CVXPY.
SCS uses fixed-point iterations to solve cone programs in standard form:
\begin{equation}\label{eq:cone_problem}
  \minimize\; c^{T}x\;\; \subjectto \;\; Ax + s = b, \;\;\;\; (x,s) \in \mathbb{R}^n \times \mathcal{K},
\end{equation}
where $x \in \mathbb{R}^n$ is the primal variable, $s \in
\mathbb{R}^m$ is the primal slack variable, $y \in \mathbb{R}^m$ is
the dual variable, and $r \in \mathbb{R}^n$ is the dual residual. The
set $\mathcal{K} \in \mathbb{R}^m$ is a non-empty convex cone.
The fixed-point computations in SCS consists of
a projection onto an affine subspace by solving a linear system
followed by a projection onto the convex cone constraints.

\subsection{Designing Neural SCS}\label{sec:scs_neural_alg}
We now describe how we design Neural SCS as a realization of
\Cref{alg:neural-fp} in three key steps: modeling, differentiating
through SCS, and designing the objective.

\paragraph{Modeling.}
The input parameters $\theta$ come from the initializations of the
neural networks that we train, $\initnet$ and $\accnet$.  To construct
the input context $\phi$ for a problem instance, we convert the
problem instance into its standard form (\cref{eq:cone_problem}), and
use the quantities $A, b$ and $c$, \ie $\phi=[v(A); b; c]$ where $v:
\mathbb{R}^{m \times n} \rightarrow \mathbb{R}^{mn}$ vectorizes the
matrix $A$. We use an MLP for $\initnet$, and a multi-layer LSTM or
GRU for $\accnet$.

\paragraph{Differentiating through SCS.}
Optimizing the loss in \cref{eq:obj} requires that we differentiate
through the fixed-point iterations of SCS:
1) For the {\em linear system solve.} We use implicit differentiation, \eg as
described in \cite{barron2016fast}. Further, for differentiating
through SCS, for a linear system $Qu = v$, we only need to obtain
the derivative $\frac{\partial u}{\partial v}$, since the
fixed-point computation repeatedly solves linear systems with the
same $Q$, but different $v$. This also lets us use an LU
decomposition of $Q$ to speed up the computation of the original
linear system solve and its derivative.
2) for the {\em cone projections,} we use the derivatives
from \citet{ali2017semismooth,busseti2019solution}.

\begin{table}[t]
  \centering
  \caption{Sizes of convex cone problems in standard form}
  \resizebox{0.75\textwidth}{!}{\begin{tabular}{llll}
    \toprule
    & \textit{Lasso} & \textit{PCA} & \textit{Kalman Filter} \\ \midrule
    Variables $n$ & 102 & 741 & 655 \\
    Constraints $m$ & 204 & 832 & 852 \\
    nonzeros in $A$ & 5204 & 1191 & 1652 \\
    \bottomrule
    \end{tabular}\hspace{5mm}
  \begin{tabular}{lllll}
    \toprule
    && \textit{Lasso} & \textit{PCA} & \textit{Kalman Filter} \\ \midrule
    \multirow{4}{*}{\rotatebox[origin=c]{90}{Cone dims}} &
    Zero & 0 & 90 & 350 \\
    & Non-negative & 100 & 181 & 100\\
    & PSD & none & [33] & none \\
    & Second-order & [101, 3] & none & [102] + [3]$\times$100 \\
    \bottomrule
  \end{tabular}}
  \label{table:prob_data}
\end{table}

\paragraph{Designing the Loss.}
The natural choice for the learning objective is the fixed-point
residual norm of SCS. With this
objective, the interacting algorithmic components of SCS
cause $\accnet$ and $\initnet$ to learn poor models for the cone
problem. In particular, SCS scales the iterates of feasible problems
by $\tau$ for better conditioning.  However, this causes a serious
issue when optimizing the fixed-point residuals: shrinking the
iterate-scaling $\tau$ artificially decreases the fixed-point
residuals, allowing $\accnet$ to have a good loss even with poor
solutions.

We eliminate this issue by normalizing each $x_t$ by its
corresponding $\tau$, similar to \citet{busseti2019solution}.
Thus, the fixed-point residual norm becomes the
$||x_t/\tau_t - f(x_t, \phi)/\tau_{f(x_t, \phi)}||$. We are then
always measuring the residual norm with $\tau=1$ for the learning
objective, which does not modify the cone program that we
are optimizing
In addition, with this objective, we no longer need to
learn or predict from $\tau$ in the models $\initnet$ and $\accnet$.

\subsection{Experiments} \label{sec:scs_prob_desc}\label{sec:cone_expt}

We demonstrate the experimental performance of SCS+Neural on 3 cone
problems: Lasso (\cite{tibshirani1996lasso}), Robust PCA
(\cite{candes2011robustpca}) and Robust Kalman Filtering, chosen similarly to
\cite{o2016conic}. \Cref{table:prob_data} summarizes problem sizes,
types of cones, and cone sizes used in our experiments. We use Adam
\citep{kingma2014adam} to train for 100,000 model updates. We perform
a hyperparameter sweep, and select models with the best validation
loss in each problem class. For SCS+AA, we use the default
history of 10 iterations. \Cref{sec:expt_addl} describes
additional training details and the source code for our
experiments is available online at
\href{https://github.com/facebookresearch/neural-scs}{github.com/facebookresearch/neural-scs}.

\begin{figure*}[ht]
  \vspace{-5mm}
  \centering
  \begin{subfigure}{0.99\textwidth}
    \includegraphics[width=.99\linewidth]{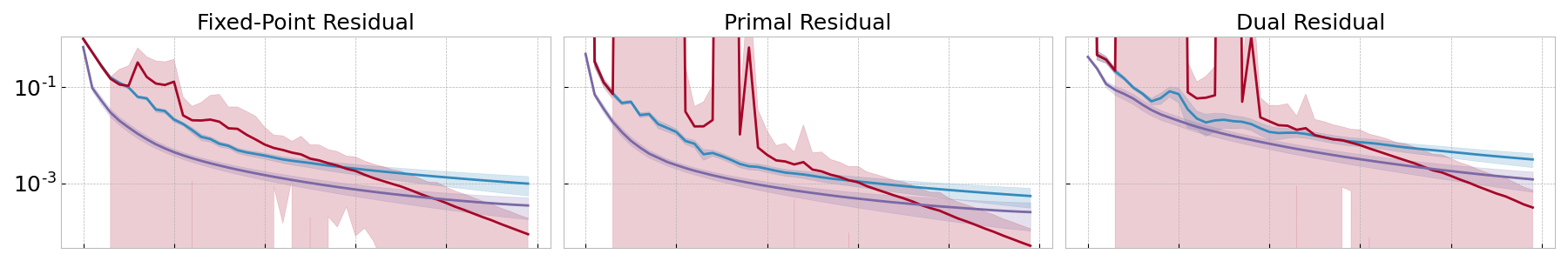}
    \caption{Lasso}\label{fig:main_lasso}
  \end{subfigure}
  \begin{subfigure}{0.99\textwidth}
    \includegraphics[width=0.99\linewidth]{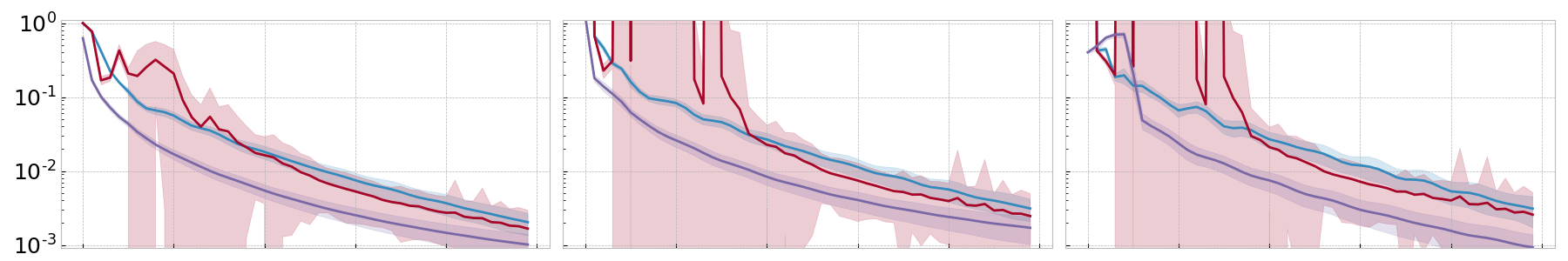}
    \caption{Robust PCA}\label{fig:main_pca}
  \end{subfigure}
  \begin{subfigure}{0.99\textwidth}
    \includegraphics[width=0.99\linewidth]{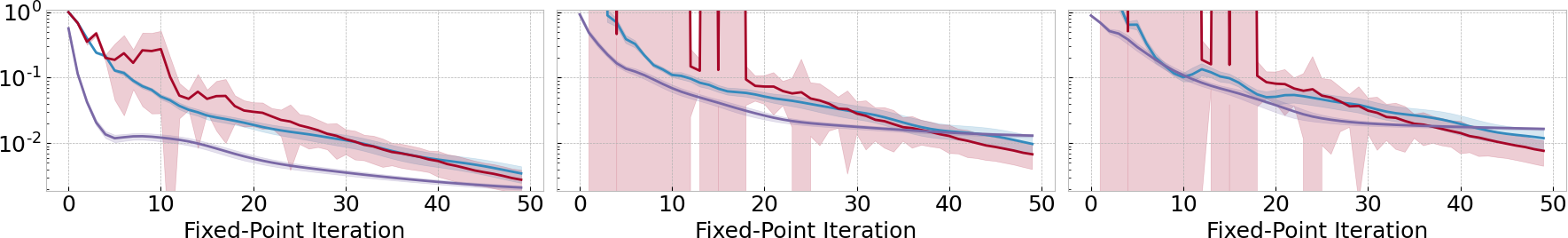}
    \caption{Robust Kalman Filtering}\label{fig:main_kf}
  \end{subfigure} \\[4mm]
  \cblock{52}{138}{189}\hspace{1.5mm}SCS\hspace{1.5mm}
  \cblock{166}{6}{40}\hspace{1.5mm}SCS+AA\hspace{1.5mm}
  \cblock{128}{114}{179}\hspace{1.5mm}SCS+Neural
  \caption{Neural accelerated SCS: Lasso, Robust PCA and Robust Kalman
    filtering}
  \label{fig:scs_neural_main}
  \vspace{-10pt}
\end{figure*}

\paragraph{Results.}
As an initial proof-of-concept, our experimental results focus on the
number of iterations required to achieve required accuracy with
SCS+Neural. \Cref{fig:scs_neural_main} shows the fixed-point, primal
and dual residuals for SCS, SCS+AA, and SCS+Neural. It shows the mean
and standard deviation of each residual per iteration, aggregated over
all test instances for each solver. SCS+Neural consistently
reaches a lower residual much faster than SCS or SCS+AA. e.g., in
Lasso (\cref{fig:main_lasso}) SCS+Neural reaches a fixed-point
residual of 0.001 in 25 iterations, while SCS+AA and SCS
take 35 and 50 iterations and SCS respectively. Our improvement
for Kalman filtering (\cref{fig:main_kf}) is even higher: we
reach a fixed-point residual of 0.01 in 5 iterations, compared to the
30 iterations taken by SCS and SCS+AA.  In addition, SCS+AA
consistently has high standard deviation, due to its well-known
stability issues.

Improving the fixed-point residuals earlier also results in
improving the primal/dual residuals earlier. For
Robust PCA (\cref{fig:main_pca}), this improvement lasts throughout
the 50 iterations. However, SCS+AA has a slight
edge in the later iterations
for Lasso and Kalman filtering, especially in the primal/dual residuals.
These can be improved by adding
a regularizer with the final primal-dual residuals to the loss
(discussed in \Cref{sec:res_ablations}).

\begin{figure}[t]
  \centering
  \includegraphics[width=0.97\textwidth]{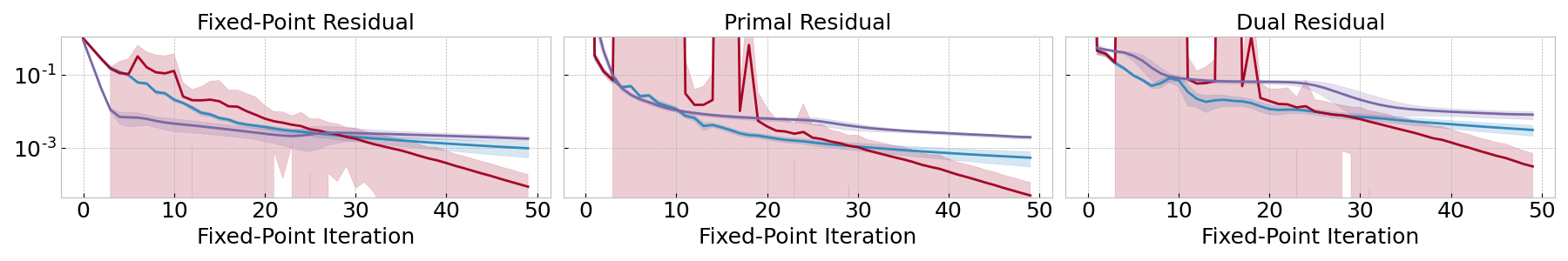}
  \cblock{52}{138}{189}\hspace{1.5mm}SCS\hspace{1.5mm}
  \cblock{166}{6}{40}\hspace{1.5mm}SCS+AA\hspace{1.5mm}
  \cblock{128}{114}{179}\hspace{1.5mm}SCS+Neural {\small (No $\tau$ normalization)}
  \caption{Lasso without $\tau$ normalization: a failure mode of
    neural acceleration (that SCS+Neural overcomes with design).}
  \label{fig:tau_norm_full}
  \vspace{-10pt}
\end{figure}

{\em Importance of $\tau$ Normalization in Objective.}
\Cref{fig:tau_norm_full} shows the residuals obtained for Lasso when
SCS+Neural does not use $\tau$ normalization in the objective. The
primal/dual residuals are worse than SCS and SCS+AA.
The fixed-point residual shows an initial improvement, but finishes worse.
As discussed in \Cref{sec:cone_prog}, this
happens when SCS+Neural achieves a low loss by simply learning a low
$\tau$, which we show in \cref{sec:tau_verify}.

\section{Conclusion and future directions}
We have demonstrated learned fixed-point acceleration for convex optimization.
Future directions include scaling to larger convex optimization
problems and accelerating fixed-point
iterations in other domains, such as
in motion planning \citep{mukadam2016gaussian},
optimal transport \citep{mena2018learning}, and
deep equilibrium models \citep{bai2019deep,bai2020multiscale}.

\subsection*{Acknowledgments}
We thank
Akshay Agrawal,
Shaojie Bai,
Shane Barratt,
Nicola Bastianello,
Priya Donti,
Zico Kolter,
Christian Kroer,
Maximilian Nickel,
Alex Peysakhovich,
Bartolomeo Stellato,
and Mary Williamson
for insightful discussions
and acknowledge the Python community
\citep{van1995python,oliphant2007python}
for developing
the core set of tools that enabled this work, including
PyTorch \citep{paszke2019pytorch},
Hydra \citep{Yadan2019Hydra},
Jupyter \citep{kluyver2016jupyter},
Matplotlib \citep{hunter2007matplotlib},
seaborn \citep{seaborn},
numpy \citep{oliphant2006guide,van2011numpy},
pandas \citep{mckinney2012python}, and
SciPy \citep{jones2014scipy}.

{\small
\bibliography{refs}}

\newpage
\appendix
\title{Neural Fixed-Point Acceleration: Supplementary
  Material}

\section{Batched and Differentiable SCS}
In this section, describe our batched and differentiable PyTorch
implementation of SCS that enables the neural fixed-point acceleration
as a software contribution.

We have implemented SCS in PyTorch, with support for the zero,
non-negative, second-order, and positive semi-definite cones. Because
our goal is to learn on multiple problem instances, we support batched
version of SCS, so that we can solve a number of problem instances
simultaneously. For this, we developed custom cone projection
operators in PyTorch that allow us to perform batched differentiation.

SCS includes a number of enhancements in order to improve its speed
and stability over a wide range of applications. Our implementation in
PyTorch supports all enhancements that improve convergence, including
scaling the problem data so that it is equilibrated, over-relaxation,
and scaling the iterates between each fixed point iteration. Our
implementation is thus fully-featured in its ability to achieve
convergence using only as many fixed point iterations as SCS.

We are also able to achieve significant improvements in speed through
the use of PyTorch JIT and a GPU. However, the focus of this work is
on proof-of-concept of neural fixed-point acceleration, and so we have
not yet optimized PyTorch-SCS for speed and scale. Our key limitation
comes from the necessity of using dense operations in PyTorch, because
PyTorch's functionality is primarily centered on dense tensors.  While
the cone programs are extremely sparse, we are unable to take
advantage of its sparsity; this limits the scale of the problems that
can be solved. We plan to address these limitations in a future
implementation of a differentiable cone solver.

\section{Application: ISTA for elastic net regression}
\label{sec:ista}
As a first simple application for demonstrating and grounding
our fixed-point acceleration, we consider the elastic
net regression setting that \citet{zhang2020globally}
uses to demonstrate the improved convergence of their
Anderson Acceleration variant.
This setting involves solving elastic net regression
\citep{zou2005regularization}
problems of the form
\begin{equation}
  \minimize \frac{1}{2}||Ax-b||_2^2 +
    \mu\left(\frac{1-\beta}{2}||x||_2^2 + \beta||x||_1\right),
  \label{eq:enr}
\end{equation}
where $A\in\R^{m\times n}$, $b\in\R^m$.
We refer to the objective here as $g(x)$.
We solve this with the fixed point computations from the
iterative shrinkage-thresholding algorithm
\begin{equation}
  f(x) = S_{\alpha\mu/2}\left(
    x-\alpha\left(A^\top(Ax-b)+\frac{\mu}{2}x\right)\right),
  \label{eq:ista}
\end{equation}
with the shrinkage operator $S_\kappa(x)_i = {\rm
  sign}(x_i)(|x_i|-\kappa)_+$.  We follow the hyper-parameters and
sampling procedures described in \citet{zhang2020globally}
and use their Anderson Acceleration with a
lookback history of 5 iterations.  We set $\mu=0.001\mu_{\rm max}$,
$\mu_{\rm max}=||A^\top b||_\infty$, $\alpha=1.8/L$, $L=(A^\top A)+
\mu/2$, and $\beta=1/2$.  We take $m=n=25$ and sample $A$ from a
Gaussian, $\hat x$ from a sparse Gaussian with sparsity 0.1, and
generate $b=A{\hat x} + 0.1w$, where $w$ is also sampled from a
Gaussian.

\begin{figure}
  \centering
  \includegraphics[width=0.8\textwidth]{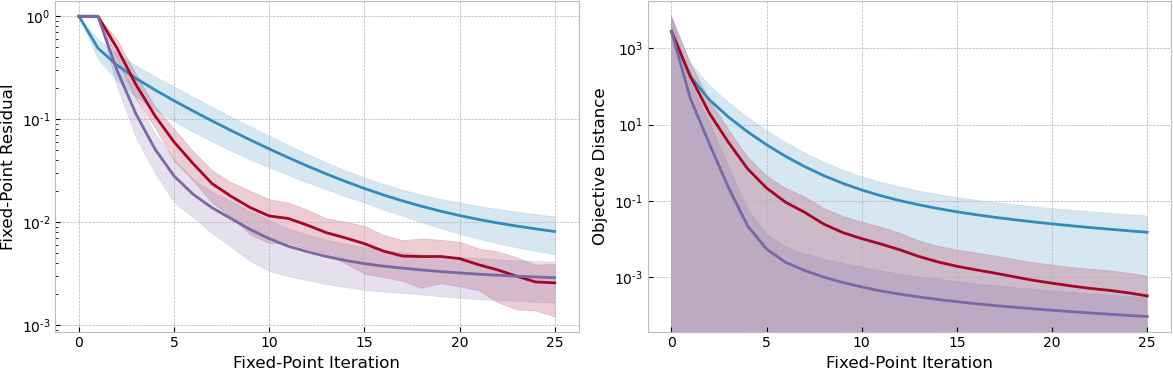} \\
    \cblock{52}{138}{189}\hspace{1.5mm}ISTA\hspace{1.5mm}
    \cblock{166}{6}{40}\hspace{1.5mm}ISTA+AA\hspace{1.5mm}
    \cblock{128}{114}{179}\hspace{1.5mm}ISTA+Neural
    \caption{Learning to accelerate ISTA for solving elastic net regression problems.
      This shows the average fixed-point residual and distance to the
      optimal objective of a single training run averaged over 50 test
      samples.}
  \label{fig:enr}
\end{figure}

We demonstrate in \cref{fig:enr} that we
competitively accelerate these fixed-point computations.
We do this using an MLP for the initialization,
GRU for the recurrent unit.
In addition to showing the training objective of the normalized
fixed-point residuals $\gR(x)/\gR_0$,
we also report the distance from the optimal objective
$||g(x)-g(x^\star)||_2^2$, where we obtain $x^\star$ by
using SCS to obtain a high-accuracy solution to \cref{eq:ista}.

\section{Additional Experiments on Neural SCS}
\subsection{Background and setup}
In this section, we provide experimental setup details for results
with SCS+Neural. We describe first the different cone problems we use,
and then describe additional experimental setup details.

\subsubsection{Cone Programs}\label{sec:cone_prob_desc}

\paragraph{Lasso.}
Lasso \cite{tibshirani1996lasso} is a well-known machine learning
problem formulated as follows:
$$\minimize_{z}\; (1/2) ||Fz - g||_2^2 +
\mu||z||_1$$
where $z \in \mathbb{R}^p$, and where
$F \in \mathbb{R}^{q \times p}$, $g \in
\mathbb{R}^p$ and $\mu \in \mathbb{R}_+$ are data. In our experiments,
we draw problem instances from the same distributions as
\cite{o2016conic}: we generate $F$ as $q \times p$ matrix with entries
from $\mathcal{N}(0,1)$; we then generate a sparse vector $z^*$ with
entries from $\mathcal{N}(0,1)$, and set a random 90\% of its entries
to 0; we compute $g = Fz^* + w$, where $w \sim \mathcal{N}(0,0.1)$; we
set $\mu = 0.1||F^{T}g||_\infty$. We use $p = 100$ and $q = 50$.

\paragraph{Robust PCA.}
Robust Principal Components Analysis~\cite{candes2011robustpca}
recovers a low rank matrix of measurements that have been corrupted
by sparse noise by solving
\begin{alignat}{2}
  & \minimize\; && ||L||_* \nonumber \\
  & \subjectto && ||S||_1 \leq \mu \nonumber \\
  & && L + S = M \nonumber
\end{alignat}
where variable $L \in \mathbb{R}^{p \times q}$ is the original
low-rank matrix, variable $S \in \mathbb{R}^{p \times q}$ is the noise
matrix, and the data is $M \in \mathbb{R}^{p \times q}$ the matrix of
measurements, and $\mu \in \mathbb{R}_+$ that constrains the corrupting
noise term.

Again, we draw problem instances from the same distributions as
~\cite{o2016conic}: we generate a random rank-$r$ matrix $L^*$, and a
random sparse matrix $S^*$ with no more than $10\%$ non-zero
entries. We set $\mu = ||S^*||_1$, and $M = L^* + S^*$. We use $p =
30$, $q = 3$ and $r = 2$.

\paragraph{Robust Kalman Filtering.} Our third example applies robust
Kalman filtering to the problem of tracking a moving vehicle from
noisy location data. We follow the modeling of
\cite{cvxpy:kalmanfilter} as a linear dynamical system. To describe
the problem, we introduce some notation: let $x_t \in \mathbb{R}^n$
denote the state at time $t \in \{0 \ldots T-1\}$, and
$y_t \in \mathbb{R}^r$ be the state measurement The dynamics of the
system are denoted by matrices: $A$ as the drift matrix, $B$ as the
input matrix and $C$ the observation matrix. We also allow for noise
$v_t \in \mathbb{R}^r$, and input to the dynamical system
$w_t \in \mathbb{R}^m$. With this, the problem model becomes:
\begin{alignat}{3}
  & \operatorname{minimize} \quad && \Sigma_{t=0}^{N-1}(||w||_2^2 + \mu\psi_\rho(v_t)) \nonumber \\
  & \operatorname{s.t.} && x_{t+1} = Ax_{t} + Bw_{t}, && t \in [0 \ldots T-1] \nonumber \\
  & && y_t = Cx_t + v_t, && t \in [0 \ldots T-1] \nonumber
\end{alignat}
where our goal is to recover $x_t$ for all $t$, and where
$\psi_\rho$ is the Huber function:
\begin{equation*}
\psi_\rho(a) =
\begin{cases}
||a||_2 & ||a||_2 \leq \rho \\
2 \rho ||a||_2 - \rho^2 & ||a||_2 \geq \rho
\end{cases}
\end{equation*}

We set up our dynamics matrices as in \cite{cvxpy:kalmanfilter}, with
$n=50$ and $T=12$. We generate $w^*_t \sim \mathcal{N}(0,1)$, and
initialize $x^*_0$ to be $\mathbf{0}$, and set $\mu$ and $\rho$ both
to $2$. We also generate noise $v^*_t \sim \mathcal{N}(0,1)$, but
increase $v^*_t$ by a factor of 20 for a randomly selected $20\%$ time
intervals. We simulate the system forward in time to obtain $x^*_t$
and $y_t$ for $T$ time steps. Table~\ref{table:prob_data}
summarizes the problem instances.

\subsubsection{Experimental Setup: Additional Details}\label{sec:expt_addl}
For each problem, we create a training set of 100,000 problem
instances (50,000 for Kalman filtering), and validation and test sets
of 512 problem instances each (500 for Kalman filtering). We allow
each problem instance to perform 50 fixed-point iterations for both
training and evaluation. We perform a hyperparameter sweep across
the parameters of the model, Adam, and training setup, which we
detail in Table~\ref{table:sweep}.

\begin{table}[!ht]
  \centering
  \caption{Parameters used for hyperparameter sweep of SCS+Neural}
  \label{table:sweep}
  \resizebox{0.5\textwidth}{!}{\begin{tabular}{ll} \toprule
    \multicolumn{2}{c}{Adam} \\ \midrule
    learning rate & [$10^{-4}$, $10^{-2}$] \\
    $\beta_1$ & 0.1, 0.5, 0.7, 0.9 \\
v    $\beta_2$ & 0.1, 0.5, 0.7, 0.9, 0.99, 0.999 \\
    cosine learning rate decay & True, False \\
    \midrule \multicolumn{2}{c}{Neural Model} \\ \midrule
    - use initial hidden state & True, False \\
    - use initial iterate & True, False \\
    Initializer: & \\
    - hidden units & 128, 256, 512, 1024 \\
    - activation function & relu, tanh, elu \\
    - depth & [0 \ldots 4] \\
    Encoder: & \\
    - hidden units & 128, 256, 512, 1024 \\
    - activation function & relu, tanh, elu \\
    - depth & [0 \ldots 4] \\
    Decoder: & \\
    - hidden units & 128, 256, 512, 1024 \\
    - activation function & relu, tanh, elu \\
    - depth & [0 \ldots 4] \\
    - weight scaling & [2.0, 128.0] \\
    Recurrent Cell: & \\
    - model & LSTM, GRU \\
    - hidden units & 128, 256, 512, 1024 \\
    - depth & [1 \ldots 4] \\
    \midrule \multicolumn{2}{c}{Misc} \\ \midrule
    max gradient for clipping & [10.0, 100.0] \\
    batch size & 16, 32, 64, 128 [Lasso \& PCA] \\
    & 5, 10, 25, 50 [Kalman filter] \\ \bottomrule
  \end{tabular}}
\end{table}

\subsection{Additional Results.}

\subsubsection{Ablations}

We present ablations that highlight the importance of the different
pieces of SCS+Neural, using Lasso as a case study.

\begin{figure}[t]
  \centering
  \includegraphics[width=0.5\textwidth]{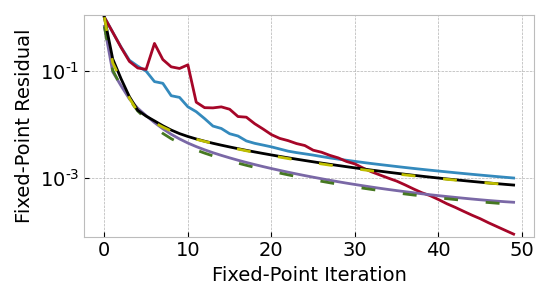} \\
      \cblock{52}{138}{189}\hspace{1.5mm}SCS\hspace{1.5mm}
  \cblock{166}{6}{40}\hspace{1.5mm}SCS+AA\hspace{1.5mm} \\
  SCS+Neural (\cblock{3}{3}{3}\hspace{1.5mm} None \hspace{1.5mm}
  \cblock{193}{193}{8}\hspace{1.5mm} Hidden \hspace{1.5mm}
  \cblock{92}{136}{60}\hspace{1.5mm} Iterate \hspace{1.5mm}
  \cblock{128}{114}{179}\hspace{1.5mm} Both)
  \caption{Initializer ablations: Lasso}
  \label{fig:lasso_initializer}
\end{figure}

\noindent
{\em Initializer.} Our first ablation examines the importance of the
learned initializer $\initnet(\phi)$ and the initial iterate
and hidden state that it provides. We modify $\initnet$ to
output four possibilities: (1) neither initial iterate nor hidden
state, (2) only the initial hidden state $h_1$, (3) only the initial
iterate $x_1$, and (4) both the initial iterate and hidden state
$[x_1, h_1]$. Note that in Case (1), the initial context $\phi$ is not
used by the neural acceleration, while Case (4) matches
\cref{alg:neural-fp}.

\Cref{fig:lasso_initializer} shows the results for the four cases of
$\initnet$ in SCS+Neural, along with SCS and SCS+AA for
comparison. They show the mean of all the test instances per
iteration, averaged across three runs with different seeds.  First,
all four cases of SCS+Neural improve significantly over SCS and SCS+AA
in the first 10 iterations, and are near-identical for the first 5-7
iterations. Further, two of the cases, i.e., Case (1) (where
$\initnet$ does not output anything), and Case (2) (where it only
outputs $h_1$) show significantly less improvement than the other two
cases; they are also near-identical. In addition, Case (3) (where
$\initnet$ outputs just the initial iterate $x_1$) is also
near-identical to Case (4) (where it outputs both $[x_1, h_1]$). This
suggests that $\initnet$ able to start the fixed-point iteration with
a good $x_1$, while the initial $h_1$ it has learned does not have
much impact.

\subsubsection{Regularizing with Primal-Dual Residuals.}\label{sec:res_ablations}
We can also optimize other losses beyond the fixed-point residuals
to reflect more properties that we want our fixed-point solutions
to have.
Here we discuss how we can add the primal and dual residuals to the
loss, which are different quantities that the fixed-point residuals.
The loss is designed to minimize the fixed-point residual as early as possible,
so sometimes, we see that the final primal-dual residuals of SCS+Neural
are slightly worse than SCS and SCS+AA.

Because the primal/dual residuals also converge under the fixed-point
map, we can adapt the loss to include them primal/dual residuals as
well, \ie, similar to \cref{eq:obj}, we can define an updated
learning objective:
\begin{equation}
  \minimize_\theta\; \E_{\phi\sim \gP(\phi)} (1 - \lambda) \sum_{t<T}
  \gR(x_t; \phi) / \gR_0(\phi) + \lambda ||[p(x_T, \phi);d(x_T, \phi)]||_2
  \label{eq:reg_obj}
\end{equation}
where $\lambda \in [0, 1]$ is the regularization parameter, $p$ and
$d$ are the primal and dual residuals at $x_T$. At $\lambda=0$, this
is our original objective \cref{eq:obj}; at $\lambda=1$, this
objective ignores the fixed-point residuals and only minimizes the
final primal and dual residuals obtained after $T$ iterations.
We ablate $\lambda$ in our experiments.

\begin{figure*}[t]
  \centering
  \includegraphics[width=\textwidth]{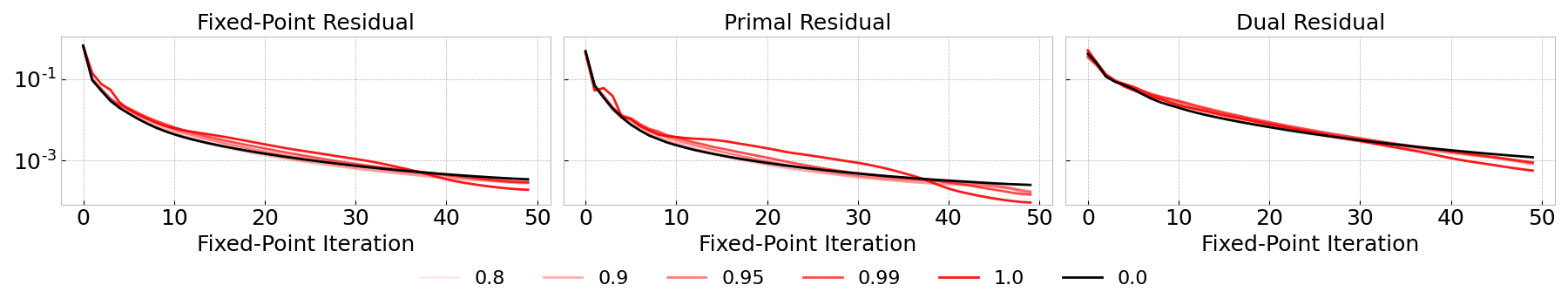}
  \caption{Ablations for regularization with primal \& dual residuals: Lasso}
  \label{fig:lasso_regularize}
\end{figure*}

Our next ablation examines the impact of regularizing the loss with
the final primal/dual residuals. \Cref{fig:lasso_regularize} shows all
three residuals for SCS+Neural for $\lambda$ ranging from 0.8 to 1.0,
in addition to the original SCS+Neural (with $\lambda =
0$). We only focus on high $\lambda$ because we see only
marginal differences from the original SCS+Neural at lower $\lambda$.
For clarity, we show only the means over all test instances for all
seeds; the standard deviations are similar to the earlier Lasso
experiments. As $\lambda$ increases, all three residuals
get a little worse than the original SCS+Neural in early iterations,
while there is an improvement in all three residuals in the later
iterations (past iteration 35). The maximum improvement in the final
primal and dual residuals at $\lambda=1$, when the learning objective
is to minimize only the final primal/dual residuals. These results
suggest that this regularization could be used to provide a flexible
tradeoff of the residuals of the final solution for the speed of
convergence of the fixed-point iteration.

\subsubsection{$\tau$ Normalization}\label{sec:tau_verify}
\begin{figure}[t]
  \centering
  \includegraphics[width=0.5\textwidth]{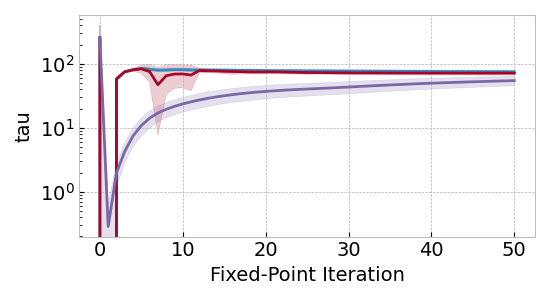} \\
  \cblock{52}{138}{189}\hspace{1.5mm}SCS\hspace{1.5mm}
  \cblock{166}{6}{40}\hspace{1.5mm}SCS+AA\hspace{1.5mm}
  \cblock{128}{114}{179}\hspace{1.5mm}SCS+Neural {\small (No $\tau$
    normalization)}
  \caption{We observe that without $\tau$ normalization,
    a failure mode of neural acceleration is that it
    learns to produce low $\tau$ values that
    artificially reduce the fixed-point residuals
    while not nicely solving the optimization problem.}
  \label{fig:example_tau_values}
  \vspace{-10pt}
\end{figure}

We can understand the behavior of $\accnet$ by examining how $\tau$
changes over the fixed-point iterations.
\Cref{fig:example_tau_values} shows the mean and standard deviation of
the learned $\tau$ values, averaged across all test instances and
across runs with all seeds. Note that SCS and SCS+AA quickly find
their $\tau$ (by iteration 3-4), and deviate very little from
it. SCS+Neural, however, starts at a very low $\tau$ that slowly
increases; this results in very low initial fixed-point residuals (and
thus a better loss for $\accnet$), but poor quality solutions with
high primal/dual residuals.

\subsubsection{Visualizing Convergence}\label{sec:visual}
Lastly, we discuss in more detail the visualizations of
convergence that we illustrated in \Cref{fig:visual}.
\Cref{fig:kalman_samples} shows the solutions of two different test
instances for Robust Kalman filtering at iterations 5, 10, 20 and
50. Lighter paths show earlier iterations, and darker paths show later
iterations. For both instances, e see that SCS+Neural has few visible
light (intermediate) paths; most of them are covered by the final dark
path, and those that are visible are of the lightest shade.  This
indicates that SCS+Neural has mostly converged by iteration 5, unlike
SCS and SCS+AA, which have many more visible intermediate (light)
paths around their final path.
\Cref{fig:pca_samples} shows the solutions for two instances in Robust
PCA at iterations 10, 15 and 20 for SCS, SCS+AA and SCS+Neural. It is
clear that, for both instances, the SCS+Neural has almost converged by
iteration 10. In contrast, SCS and SCS+AA show many more visible
distinct points at iterations 10 and 15, indicating they have not yet
converged.

\end{document}

%% file: math_commands.tex
\usepackage{amsmath,amsfonts,bm}
\usepackage{mathtools}
\usepackage{multirow}
\usepackage{tabularx}
\usepackage{graphicx}
\usepackage{adjustbox}
\usepackage{xspace}

\newcommand{\mset}[1]{\left\{\kern-.5em\left\{ #1 \right\}\kern-.5em\right\}}
\newcommand{\mmset}[1]{\{\kern-.4em\{ #1 \}\kern-.4em\}}

\def\eqref#1{equation~\ref{#1}}

\def\1{\bm{1}}

\def\vec1{{\bm{1}}}

\DeclareMathAlphabet{\mathsfit}{\encodingdefault}{\sfdefault}{m}{sl}
\SetMathAlphabet{\mathsfit}{bold}{\encodingdefault}{\sfdefault}{bx}{n}

\def\gP{{\mathcal{P}}}

\def\gR{{\mathcal{R}}}

\newcommand{\E}{\mathbb{E}}

\newcommand{\R}{\mathbb{R}}

\DeclareMathOperator*{\minimize}{minimize}

\DeclareMathOperator*{\subjectto}{subject\;to}

\newcommand{\defeq}{\vcentcolon=}